\documentclass{article}
\usepackage{spconf,amsmath,graphicx}
\graphicspath{{./images/}}
\usepackage{amsmath} 
\usepackage{amssymb}  
\usepackage{hyperref}
\hypersetup{
  colorlinks, linkcolor=red
}

\title{ACNet: Attention Based Network to Exploit Complementary Features for RGBD Semantic Segmentation}
\name{Xinxin Hu, Kailun Yang, Lei Fei and Kaiwei Wang\thanks{This work has been partially funded through the project “Research on Vision Sensor Technology Fusing Multidimensional Parameters” (111303-I21805) by Hangzhou SurImage Technology Co., Ltd and supported by Hangzhou KrVision Technology Co., Ltd (krvision.cn). The authors would like to acknowledge Juan Wang for the GPU support.}}
\address{College of Optical Science and Engineering, Zhejiang University}
\begin{document}
%
\maketitle
\begin{abstract}
Compared to RGB semantic segmentation, RGBD semantic segmentation can achieve better performance by taking depth information into consideration. However, it is still problematic for contemporary segmenters to effectively exploit RGBD information since the feature distributions of RGB and depth (D) images vary significantly in different scenes. In this paper, we propose an Attention Complementary Network (ACNet) that selectively gathers features from RGB and depth branches. The main contributions lie in the Attention Complementary Module (ACM) and the architecture with three parallel branches. More precisely, ACM is a channel attention-based module that extracts weighted features from RGB and depth branches. The architecture preserves the inference of the original RGB and depth branches, and enables the fusion branch at the same time. Based on the above structures, ACNet is capable of exploiting more high-quality features from different channels. We evaluate our model on SUN-RGBD and NYUDv2 datasets, and prove that our model outperforms state-of-the-art methods. In particular, a mIoU score of 48.3\% on NYUDv2 test set is achieved with ResNet50. We will release our source code based on PyTorch and the trained segmentation model at \url{https://github.com/anheidelonghu/ACNet}.
\end{abstract}
\section{Introduction}
\label{sec:intro}

Semantic segmentation is a basic task of computer vision, whose purpose is to partition an image into several coherent semantically-meaningful parts. Compared with traditional approaches that need to be deployed in complex separate ways, semantic segmentation can be utilized to unify diverse detection tasks desired by navigation systems, at least in standard outdoor conditions~\cite{yang2018unifyingerfpsp}\cite{yang2019can}.

 
 \begin{figure}[t]
    \centering
    \includegraphics[width=0.48\textwidth]{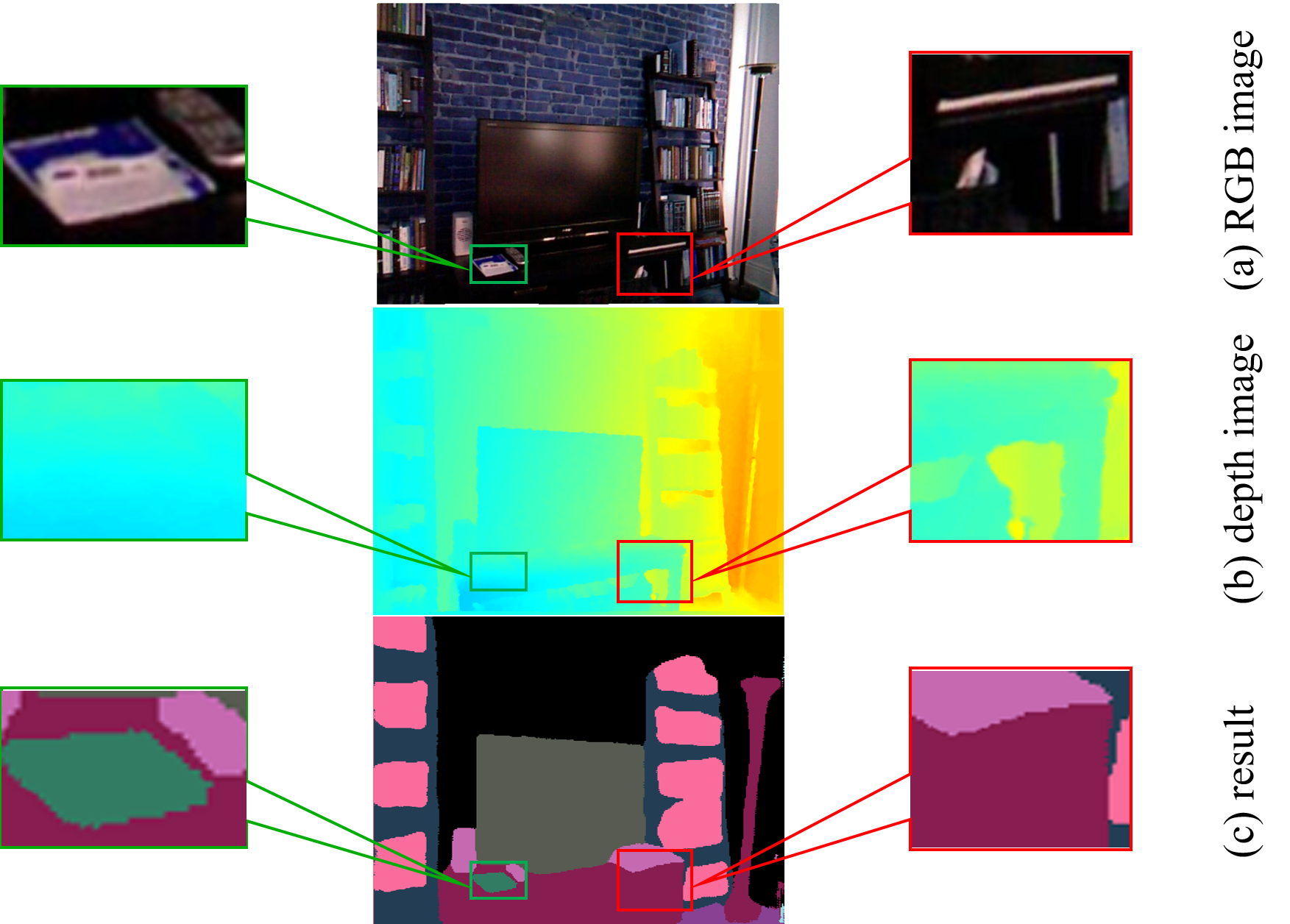}
    \vskip-2ex
    \caption{RGB and depth images have different distributions of features that are appropriately exploited by our ACNet.}
    \label{rgbdcmp}
\vskip-4ex
\end{figure}

 
In contrast, indoor semantic segmentation that has not been thoroughly investigated, remains challenging in several aspects. For example, it suffers from uneven illumination and messy spatial overlapping. With the emergence and development of RGBD cameras (such as RealSense, Kinect, Xition etc.), indoor semantic segmentation can be benefited from RGBD observations that encode real-world geometric information, which theoretically leads to better segmentation performance compared to RGB semantic segmentation. Towards this end, there were a few attempts like~\cite{gupta2014learning}\cite{he2017std2p} that treated depth image as an additional channel, and used the method similar to RGB semantic segmentation to implement RGBD semantic segmentation. In~\cite{rednet}, two neural networks branches were designed for RGB input and depth input, which were merged before upsampling. In~\cite{gupta2014learning}, depth images was decomposed into three channels, namely disparity, height and angle which were also treated as an RGB image. In~\cite{qi20173dgnn}, instead of simply using traditional convolution neural networks, novel graph-based networks were applied to excavate scene geometric information more sufficiently. In~\cite{wang2018depthaware}, traditional convolution was modified according to depth values. These networks designed for RGBD semantic segmentation have achieved break-through results. However, there are still some issues that need to be solved:





\begin{figure*}[!t]
    \centering
    \includegraphics[width=0.96\textwidth]{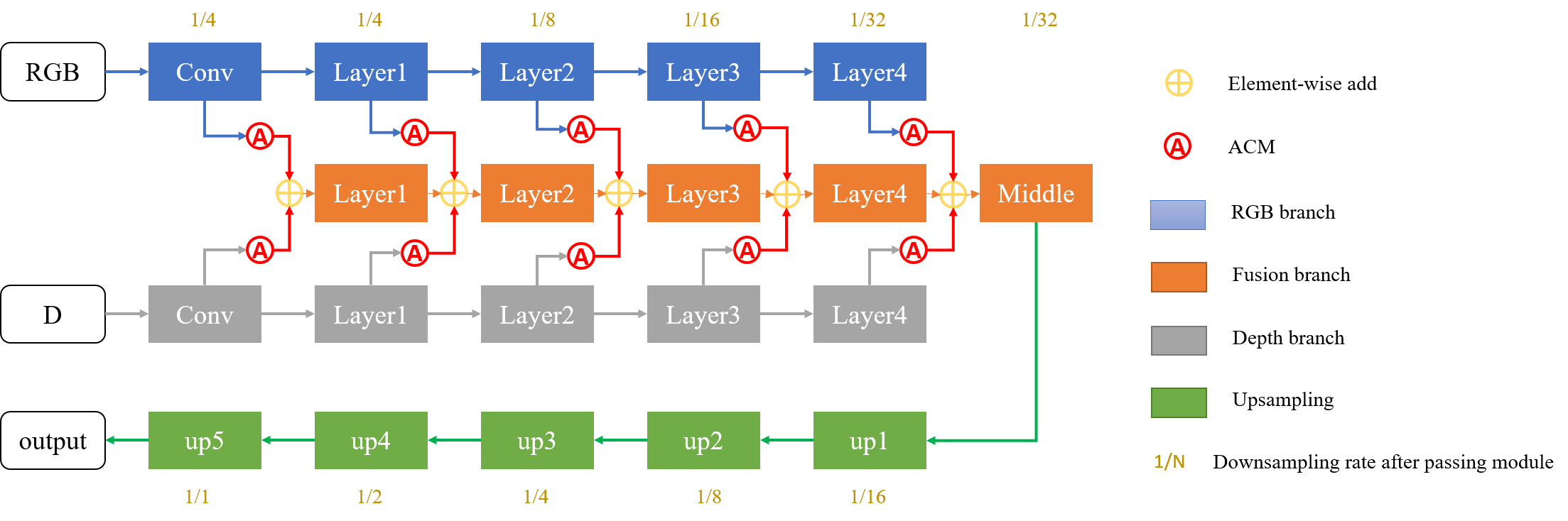}
    \vskip-3ex
    \caption{The overview of our proposed ACNet. RGB image and depth image are processed by two ResNets separately. Red arrows represent the data flow of feature maps reorganized by ACM.}
    \label{ACNet}
\vskip-2ex
\end{figure*}

\begin{itemize}

\item Although the geometric information encoded in the depth image can clearly provide additional benefits for image segmentation, the information contained in RGB image and depth image are not equivalent for each scene (shown in Fig.~\ref{rgbdcmp}). In other words, features extracted from RGB branch and depth branch by current networks may be not appropriate.
\item Conventional RGBD segmentation network can be divided into two types of architectures. One of them, such as~\cite{miou459localitysensitive}, employs two encoders to extract features from RGB and depth image respectively, and combines the features of both before or during upsampling. The other like~\cite{rednet}\cite{chen2018progressively} just fuses the RGBD features at the downsampling stage. The former can't sufficiently combine RGBD information, and the latter tends to lose original RGB and depth branches since the fusion branches take the place of them.
\end{itemize}

In this paper, we propose ACNet (shown in Fig.~\ref{ACNet}) to combine RGB and depth features by a proportion determined by the input. In ACNet, there are two independent branches based on ResNet~\cite{he2016deepresnet} to extract features for RGB and depth image separately. Several Attention Complementary Module (ACMs) are designed to obtain features from the aforementioned branches, which are determined by the amount of information they carry. There's another branch based on ResNet to process the merged features. The proposed architecture is able to keep original RGBD features flow as well as to utilize merged features in an integrated network.

\section{FRAMEWORK}
\label{sec:framework}

\begin{figure}[h]
    \centering
    \includegraphics[width=0.48\textwidth]{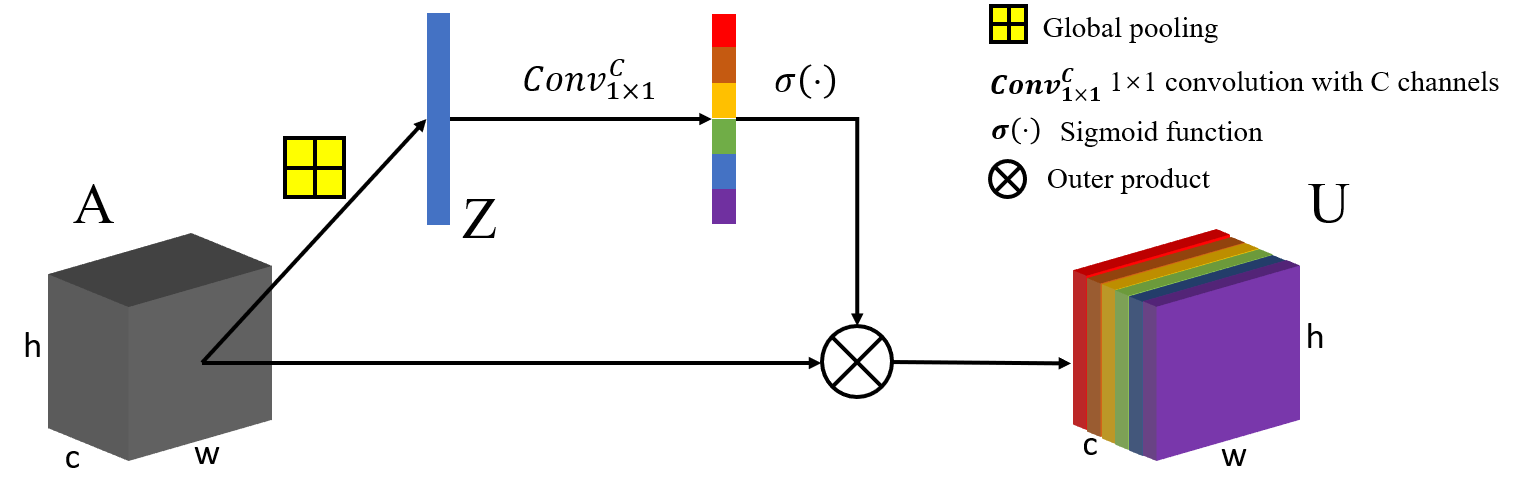}
    \vskip-3ex
    \caption{Attention Complementary Module (ACM).}
    \label{ACM}
    \vskip-2ex
\end{figure}

\textbf{Attention Complementary Module (ACM)}. As illustrated in Fig. \ref{rgbdcmp}, the information contained in RGB image and depth image vary in different regions of indoor scenes. In order to gather features selectively from RGB branch and depth branch, we have designed a set of attention modules~\cite{hu2017squeeze} to make the network focus on more informative regions. More precisely, the proposed ACM is based on channel attention~\cite{zhang2018imagechannelattention} (shown in Fig. \ref{ACM}). Assuming the input feature maps $A=[A_{1},~\cdots,~A_{C}] \in \mathbb{R}^{C \times H \times W}$, we first apply global average pooling, to have the output $Z \in \mathbb{R}^{C \times 1 \times 1}$, where $C$ denotes the number of channels, $H, W$ denote the height and width of feature maps respectively. The $k$-th~$(k\in{[1, C])}$ of $Z$ can be expressed as:

\begin{equation}
    Z_{k} = \frac{1}{H \times W} \sum_{i}^{H} \sum_{j}^{W} A_{k}(i,j)
    \label{global average pooling}
\end{equation}

Then $Z$ is reorganized by a $1 \times 1$ convolution layer with the same number of channels as $Z$. A $1 \times 1$ convolution layer is able to excavate correlations between channels, thus eliciting an appropriate weight distribution for these channels. A sigmoid function is applied to activate the convolution result, constraining the value of weight vector $V \in \mathbb{R}^{C \times 1 \times 1}$ between 0 and 1. Finally, we perform an outer product for $A$ and $V$, and the result $U \in \mathbb{R}^{C \times H \times W}$ can be expressed as:
\begin{equation}
    U = A \otimes \sigma [\phi (Z)]
\end{equation}
where $\otimes$ denotes outer product, $\sigma$ denotes sigmoid function, and $\phi$ denotes $1 \times 1$ convolution. In this way, feature maps $U$ are converted into new feature maps $U$, which contain more valid information.

\begin{figure*}[ht]
    \centering
    \includegraphics[width=0.96\textwidth]{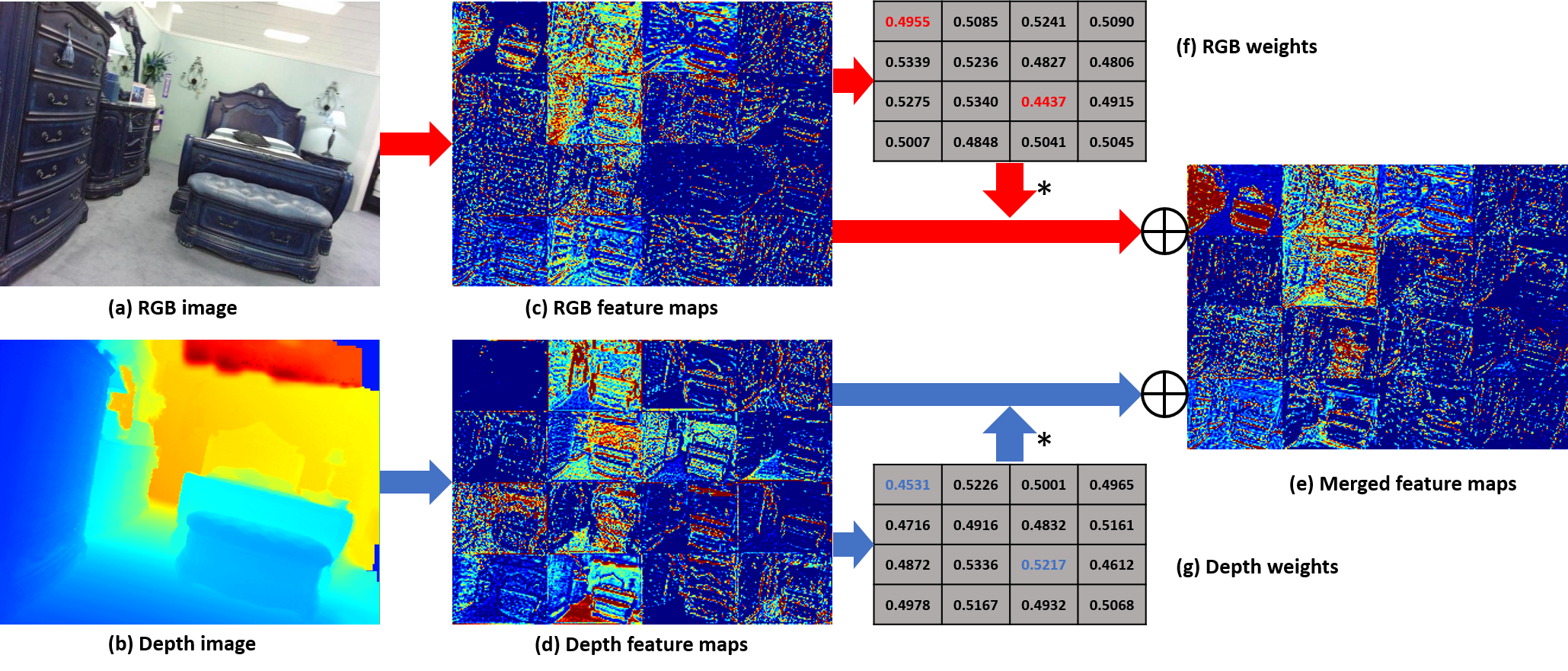}
    \vskip-3ex
    \caption{How ACM fuses complementary RGBD features into fusion branch. $*$ denotes element-wise product and $\oplus$ denotes element-wise add. The feature maps are visualized from layer2. (f) and (g) depict the weights calculated from the feature maps by ACM, which are multiplied to feature maps separately, and added into the merged features from the fusion branch.}
    \label{Analysis of ACM pic}
    \vskip-2ex
\end{figure*}

\textbf{Architecture for Feature Fusion}. A majority of state-of-the-art RGBD semantic segmentation networks use an encoder that fuses RGBD features either too early or too late~\cite{rednet}\cite{miou459localitysensitive}, which ruins the original RGB and depth information or results in a low efficiency of exploiting the carried information. In order to keep the original RGB and depth features flow during downsampling, we propose a specialized architecture for RGBD feature fusion. As illustrated in Fig.~\ref{ACNet}, two complete ResNets are deployed to extract RGB and depth features separately. Note that here the ResNet can be replaced with other networks, e.g., ERF-PSPNet~\cite{yang2019can} in efficiency-critical domains. Vitally, these two branches can preserve RGB and depth features before upsampling. After that, the fusion branch is leveraged to extract features from the merged feature maps.


\textbf{Attention Complementary Network (ACNet)}. We design an integrated network called ACNet for RGBD semantic segmentation. The backbone of ACNet is shown in Fig.~\ref{ACNet}. RGB image and depth image are inputted, and are processed by ResNet branches separately. During inference, each aforementioned branch provides a group of feature maps at every module stage, such as Conv, Layer1, etc. Then the feature maps are reorganized by ACM. After passing through Conv, the feature maps are further element-wisely added as input of fusion branch, while others are added to the output of fusion branch. In this way, both low-level and high-level features can be extracted, reorganized and fused by our ACNet. As for upsampling, we apply the skip connection like~\cite{rednet}, which appends the features in downsampling to upsampling with a quite low computation cost. 

\section{Experiments}
\label{sec:experiments}

We evaluate our method on two public datasets: 

\textbf{NYUDv2}~\cite{silberman2012indoornyu}: The NYU-Depth V2 data set (NYUDv2) contains 1,449 RGBD images with dense pixel-wise annotation. We divide the dataset into 795 training images and 654 testing images according to the official setting. We use the version with annotations on 40 classes (common ones used in the literature).

\textbf{SUN-RGBD}~\cite{song2015sun}: We use SUN-RGBD V1 which have 37 categories and contains 10,335 RGBD images with dense pixel-wise annotations, 5,285 images for training and 5,050 for testing.

As for metrics, we use the prevailing mean Intersection-over-Union over all classes (mIoU) to evaluate the performance of different semantic segmenters. 



\textbf{Implementation Details}. As for data augmentation, we apply random scaling, cropping and flipping to both RGB and depth images, and normalize them separately. For RGB images, we also randomly change their color in HSV space. For all the experiments, we use ResNet50 as the encoder, which is pre-trained on ImageNet~\cite{russakovsky2015imagenet}. Since depth image contains one channel, we average the three channels in ResNet50's first layer to one channels for the depth branch. We use focal loss~\cite{lin2018focalloss} with the focusing parameter $\gamma = 2$ to supervise the training of our network. During training stage, we also calculate the average loss of outputs from up1 to up5 in Fig.~\ref{ACNet} to better optimize our network. During testing stage, we only evaluate the metrics of last output to ensure consistency with the state of the art. We use SGD optimizer with initial learning rate 0.002, momentum 0.9 and weight decay 0.004. Batch size is set to 4 when training on one NVIDIA TITAN Xp. The learning rate is multiplied by 0.8 for every 20 iterations on SUN-RGBD and 100 iterations for NYUDv2.

\begin{figure*}[t]
    \centering
    \includegraphics[width=0.96\textwidth]{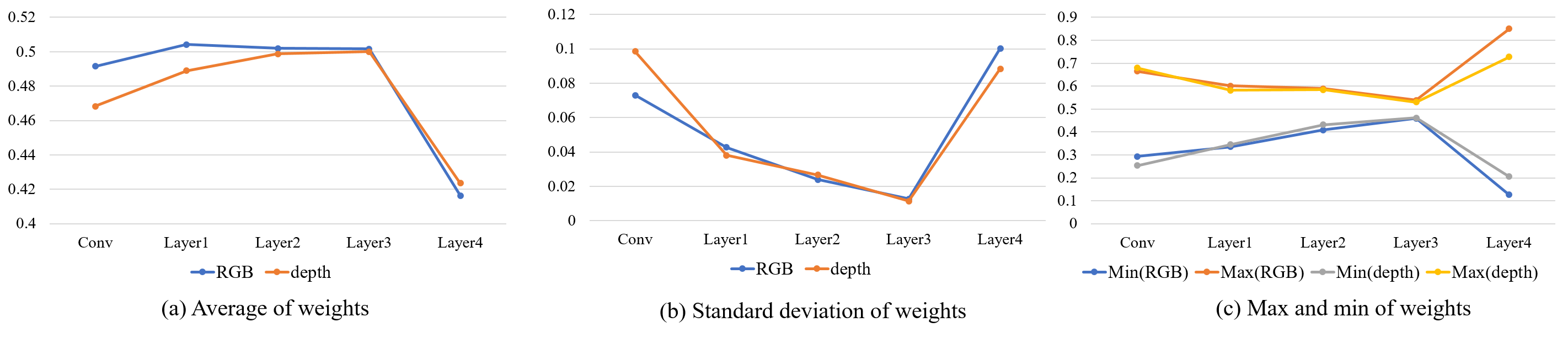}
    \vskip-3ex
    \caption{Quantitative analysis of all ACM in our model}
    \label{attention layer analysis}
    \vskip-2ex
\end{figure*}



\textbf{Analysis of the ACM}. To understand ACM better, we visualize the feature maps from layer2 (shown in Fig.~\ref{Analysis of ACM pic}) since layer2's low-level features are more consistent with visual intuitions. Note that we only visualize the first 16 of 128 feature maps for better illustration. Regarding the weights as matrices starting from (0,0), they correspond to the features maps one by one, where we can find some insightful results. At (0,0), feature map of RGB branch contains more valid information than the feature map from depth branch visually, so that ACM tends to give a higher weight to the RGB branch. In contrast, at (2,2), feature map of depth branch contains more information, therefore, depth branch gets higher weight. Finally, feature maps of the two branches are element-wisely added into feature maps of fusion branch to supplement the RGBD information.

We also evaluate weights generated by ACM at all levels (shown in Fig.~\ref{attention layer analysis}). First, we focus on averages of weights, which indicate the amount of information contained in the feature maps before inputting to ACM. 
The averages of RGB branches' weights are higher than those of depth branches in Conv and Layer1. This reveals that RGB branches always contain more valid information at lower levels since RGB images often contain many redundant textures. In addition, weights of the two branches in Layer2, Layer3 and Layer4 are queie close, which means RGB and D branches contain nearly equivalent valid information at higher levels. In addition, the average weights of Layer4 are quite low, indicating that the fusion branch might gather enough features. 
The metrics: std (standard deviation), min and max can reflect the distribution of information across channels. From Conv to Layer3, the std becomes increasingly smaller, which indicates that ACNet can homogenize the distribution of information. However, Layer4 should decide to select useful features as well as eliminate redundant features as it is the last module in the encoder. Therefore, the std of Layer4 is quite high. This experiment demonstrates that our network flattens the distribution of information across channels where complementary features are effectively exploited, which is essential for RGBD semantic segmentation.



\begin{table}[t]
\caption{Comparison with other state-of-the-art methods on NYUDv2 test set and SUN-RGBD test set.}
\vskip-1ex
\label{NYUDv2 test table}
\begin{center}
\begin{tabular}{|c|c|c|}
\hline
\textbf{Model} & \textbf{NYUDv2} & \textbf{SUN-RGBD}\\
\hline
3DGNN \cite{qi20173dgnn} & 39.9\% & 44.1\% \\
\hline
RefineNet (ResNet152) \cite{lin2017refinenet} & 46.5\% & 45.9\% \\
\hline
Depth-aware CNN \cite{wang2018depthaware} & 43.9\% & 42.0\% \\
\hline
LSD \cite{miou459localitysensitive} & 45.9\% & -\\
\hline
CFN (VGG-16) \cite{lin2017cascaded} & 41.7\% & 42.5\% \\
\hline
CFN (RefineNet-152) \cite{lin2017cascaded} & 47.7\% &\textbf{48.1\%} \\
\hline
ACNet~(ResNet-50) & \textbf{48.3\%} &\textbf{48.1\%}\\
\hline
\end{tabular}
\end{center}
\vskip-4ex
\end{table}

\textbf{Ablation Study.} To verify functionality of both ACM and the multi-branch architecture, we perform an ablation study by comparing the original model with two defective models: \textbf{Model-1} and \textbf{Model-2}. In \textbf{Model-1}, we remove all ACMs and the RGB and D branches after Conv Layer. In \textbf{Model-2}, we remove all ACMs but retain the multi-branch architecture. Our ablation study on NYUDv2 turns out that, the mIoU of \textbf{Model-1} and \textbf{Model-2} are 44.3\% and 46.8\%, verifying the multi-branch architecture and ACM lead to significant accuracy boost of 2.5\% and 1.5\%, respectively. 

\textbf{Comparison with state-of-the-art networks}. We compare our ACNet with state-of-the-art methods to prove its effectiveness. Note we adopt the most universally-used mIoU as the evaluation metric. 

Table~\ref{NYUDv2 test table} shows the result of our ACNet on NYUDv2 and SUN-RGBD test sets. The result shows that on NYUDv2, our ACNet outperforms other state-of-the-art models by 0.6\%, yielding the new record of mIoU accuracy 48.3\% on NYUDv2. On SUN-RGBD, our model~(ResNet-50) is able to reach the same mIoU as CFN~(RefineNet-152)~\cite{lin2017cascaded} by using a more lightweight backbone.



\section{CONCLUSIONS}
\label{sec:conclusions}

In this paper, we propose a novel multi-branch attention based network for RGBD semantic segmentation. The multi-branch architecture is able to gather features efficiently and doesn't destroy original RGB and depth branches' inference. The attention module can selectively gather features from RGB and depth branches according to the amount of information they contain, and complement the fusion branch by using these weighted features. Our model can resolve the problem that RGB images and depth images always contain unequal amount of information as well as different context distributions. We evaluate our model on NYUDv2 and SUN-RGBD datasets, and the experiments show that our model can outperform state-of-the-art methods.

In the future, we will explore ways to improve the real-time performance of pixel-wise image segmentation not only for RGBD semantic cognition but also for panoramic annular surrounding perception.

\newpage
\bibliographystyle{IEEEbib}
\bibliography{Template}

\end{document}